\documentclass[runningheads]{llncs}
\usepackage[T1]{fontenc}
\usepackage{graphicx}
\begin{document}
\title{MambaMorph: a Mamba-based Framework for Medical MR-CT Deformable Registration}
\titlerunning{Mamba-based Deformable Registraion}
%
\author{Tao Guo\inst{1} \and
Yinuo Wang\inst{1} \and
Shihao Shu\inst{1} \and Weimin Yuan\inst{1} \and Diansheng Chen\inst{2} \and Zhouping Tang\inst{3} \and Cai Meng\inst{1} \and
Xiangzhi Bai\inst{1}}
\authorrunning{T. Guo et al.}
%
\institute{Image Processing Center, Beihang University, Beijing, China \and Robotics Institute, Beihang University, Beijing, China \and Tongji Hospital, Huazhong University of Science and Technology, Wuhan, China
\email{\{tsai, jackybxz\}@buaa.edu.cn}}
\maketitle              
\begin{abstract}
Deformable image registration is an essential approach for medical image analysis.This paper introduces MambaMorph, an innovative multi-modality deformable registration network, specifically designed for Magnetic Resonance (MR) and Computed Tomography (CT) image alignment. MambaMorph stands out with its Mamba-based registration module and a contrastive feature learning approach, addressing the prevalent challenges in multi-modality registration. The network leverages Mamba blocks for efficient long-range modeling and high-dimensional data processing, coupled with a feature extractor that learns fine-grained features for enhanced registration accuracy. Experimental results showcase MambaMorph's superior performance over existing methods in MR-CT registration, underlining its potential in clinical applications. This work underscores the significance of feature learning in multi-modality registration and positions MambaMorph as a trailblazing solution in this field. The code for MambaMorph is available at: https://github.com/Guo-Stone/MambaMorph.

\keywords{multi-modality registration  \and Mamba \and Feature learning.}
\end{abstract}
\section{Introduction}
Deformable image registration is an essential approach for medical image analysis. Because of the surgical intervention, different imaging sequence etc., the topology of anatomical tissue in image varies a lot. Before analyzing a pair of images, it’s necessary to align them spatially via deformable image registration. For instance, there exists up to 10 mm deformation between preoperative MR and intraoperative CT due to the invasive surgery \cite{10mm}, which can’t be solved by affine registration merely. Only after registering intraoperative CT to preoperative MR, physicians are able to get the accurate anatomical brain segmentation and perform surgical navigation.

Though conventional registration approaches \cite{syn} can calculate precise and diffeomorphic displacement field, they also bring heavy computation burden and cost considerable time which are not suitable for real-time situation. In the past decade, deep learning-based registration methods, e.g. VoxelMorph 
 \cite{voxelmorph}, showcase their ability to implement quick registration whose accuracy can even be on par with their conventional counterparts. VoxelMorph is a pioneering work in deep learning-based deformable registration based on convolutional neural network (CNN), which is designed for unsupervised, weak-supervised registration and utilize scaling-and-squaring technique \cite{scaling} to integrate SVF (stationary velocity field). Inheriting from the framework of VoxelMorph, amounts of works \cite{transmorph,contrareg,gsmorph,mr-ct,synthmorph,hypermorph,r2net,cyclemorph,same} are developed for higher accuracy and multi-modality deformable registration.   

When performing multi-modality deformable registration, it may encounter several issues: lack of annotated data, hard to represent volumes of different modalities, hard to capture long-range correspondence effectively. Therefore, SynthMorph \cite{synthmorph} overcomes data deficiency by generating multi modal volumes via sampling, another work for MR-CT registration \cite{mr-ct} turns multi-modality registration to two single-modal registration via synthesis so as to bypass feature learning and TransMorph \cite{transmorph} takes the use of Swin Transformer to modeling long-range relation. To some extent, the above methods alleviate issues in multi-modality deformable registration. However, they are not highly efficient (e.g. low convergence speed, heavy computation burden) due to the poor synthetic data quality, oversized framework and quadratic complexity introduced by Transformer.

To tackle the issues mentioned above, we propose an end-to-end deformable registration algorithm empowered by contrastive learning and repropose a well-annotated MR-CT registration dataset from SynthRAD 2023 [Citation]. Specifically, we add a fine-grained feature extractor before registration module, which is trained with supervised contrastive learning loss [Citation] to extract representation better. A technique called gradient surgery [Citation] is implemented to avoid incompatibility between registration and representation learning. Besides, we make an MR-CT registration dataset based on SynthRAD 2023 dataset carefully by skull stripping, brain segmentation, intensity correction and quality control. This dataset may benefit the community which suffer from data deficiency and promote MR-CT registration research. Compared with several advanced registration algorithm, our method achieves ... in our proposed dataset.

\section{Related Work}
\subsection{Deep Learning-based Registraion}
VoxelMorph is a representative deep learning-based registration framework with competent performance and short inference time. The framework consists of a UNet backbone followed by a spatial transform network (STN) and is trained with a similarity loss and a smooth loss. Inspired by the success of VoxelMorph, amounts of following work inherit its framework and take UNet-like network as their backbone. When dealing with input, these networks treat moving volume and fixed volume as a single volume by concatenating them first. It means that feature extraction and feature matching are done by a UNet-like backbone simultaneously, which may be out of the backbone’s capacity to some extent. If the inputs are in the same modality, these frameworks may work well because the inputs share similar appearance. While the inputs’ modalities are different, it's challenging for these frameworks to extract and match feature at the same time due to the considerable difference between two inputs. It may lead to feature distortion and poor registration performance.

\subsection{Multi-modality Registration}
Volumes in different modalities (e.g. MR and CT) vary a lot in terms of intensity distribution and appearance that are difficult for current algorithms to register deformably. When it comes to cross-modal deformable registration, there are three main problems: lack of annotated data, hard to represent and hard to measure similarity. To overcome the deficiency of data, SynthMorph [Citation] generates plenty of modalities volumes and trains VoxelMorph based framework with them. It’s an ingenious method to enhance the model’s generalization to unseen modality while it converges quite slowly. [Citation] bypasses feature extraction and similarity measurement between MR and CT via image synthesis with a bulky and hard-to-train framework. XMorpher [Citation] and Cross begin to emphasize the importance of representation learning in deformable registration and introduce cross attention to exchange information between two modalities feature. Tough these methods achieve better registration performance, they don’t take use of label information sufficiently and suffer from heavy memory burden due to attention mechanism. This work reproposes a well annotated MR-CT registration dataset and presents a registration framework with a light-weighted feature extractor which is guided by anatomical semantic information.

\section{Method}
\begin{figure}
\includegraphics[width=\textwidth]{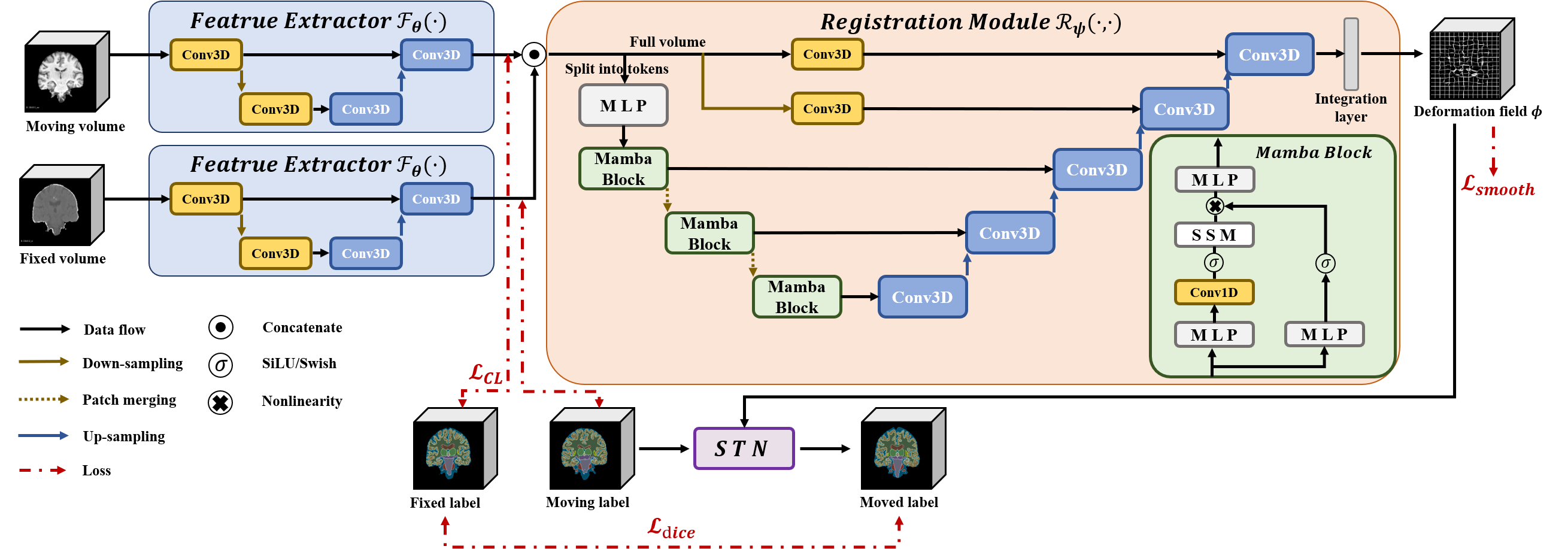}
\caption{The framework of MambaMorph} \label{fig1}
\end{figure}

To perform multi-modality (MR-CT in this paper) deformable registration, our MambaMorph adopts two proposed technique, i.e. Mamba-based registration module and contrastive feature learning. The framework of MambaMorph is illustrated in Fig.~\ref{fig1}. In addition, we repropose an MR-CT deformable registration dataset called SynthRAD Registration (SR-Reg) to alleviate the issue of well-aligned MR-CT data deficiency.

Given a pair of moving volume $x_{m}$ and fixed volume $x_{f}$, their corresponding segmentation $s_{m}$ and $s_{f}$, registration module $\mathcal{R}_{\psi}(\cdot,\cdot)$ and feature extractor $\mathcal{F}_{\theta}(\cdot)$, the registration in this paper can be formulated as follows:
\begin{equation}
\min\limits_{\psi,\theta} \mathcal{L}_{dice}(s_{m}\circ\phi,s_{f})+\lambda_{c} \mathcal{L}_{CL}(f_{z},s_{z})|_{z\in\{m,f\}}+\lambda_{s} \mathcal{L}_{smooth}(\phi)
\end{equation}
where deformation field $\phi=\mathcal{R}_{\psi}(f_{m},f_{f})$, feature $f_{z}|_{z\in\{m,f\}}=\mathcal{F}_{\theta}(x_{z})$, $\mathcal{L}_{dice}(\cdot,\cdot)$ and $\mathcal{L}_{smooth}(\cdot)$ are weak-supervised loss and smooth loss originated from VoxelMorph[C], $\mathcal{L}_{CL}(\cdot,\cdot)$ is contrastive loss.

In this section, we are about to introduce our mamba-based registration module $\mathcal{R}_{\psi}(\cdot,\cdot)$ in subsection 3.1, contrastive feature learning $\mathcal{F}_{\theta}(\cdot)$ and $\mathcal{L}_{CL}(\cdot,\cdot)$ in subsection 3.2.
Finally, we illustrate our SynthRAD Registration (SR-Reg) dataset processing approach in subsection 3.3. 

\subsection{Mamba-based registration module}
Following the framework of TransMorph[C], our MambaMorph's registration module makes some modification which is shown in Fig.~\ref{fig1}. After splitting a full size volume (e.g. 192$\times$208$\times$176 voxels in this paper), one may obtain a sequence with about 0.1 million tokens. Due to the poor long-range sequence modeling capacity and quadratic complexity w.r.t. sequence length, we assume that Transformer is not of high efficiency at this case. As a potential replacer of Transformer, Mamba is more good at dealing with considerably long sequence with nearly linear complexity. Therefore, the encoder is substituted with Mamba block instead of Swin Transformer block in our module.

Developed from SSM (State Space Model), Mamba can be seemed as an input-dependent variant of RNN. Similar to RNN and Transformer, Mamba takes as input a tensor of shape $B \times L \times C$ and gets an output of the same shape. $B$, $L$, $C$ denotes batch size, sequence length and number of channel respectively. In the main stream of Mamba block, the input are projected linearly via an MLP and sent to a 1D convolution layer then. After a SiLU/Swish layer, the tensor is processed in the SSM layer. In the SSM of Mamba, three transition matrices are generated by linear layer directly or indirectly so as to be input-dependent. Having introduced input-dependent transition matrices, the system of Mamba is not time-invariant anymore and needed to solve via recurrence instead of convolution. Another branch of Mamba introduces a gated mechanism to be selective w.r.t. data. Notably, we add sinusoidal position embedding onto the input tensor to make Mamba position-aware.

MambaMorph' registration module takes a concatenated volume as input, which is from moving volume feature and fixed volume feature. The input flows into two branches - a horizontal one with full volume and a UNet-like one where is split. At the latter branch, the volume is split into amounts of patches and seemed as a sequence of token (patch). After linear projection, consecutive stages of Mamba block and patch merging are applied on the sequence. Patch merging is a variant of down sampling, involving decreasing the amount of token from $B \times (H\times W \times D) \times C$ to $B \times (H/2\times W/2 \times D/2) \times 8C$ and projecting the number of channel from $8C$ to $2C$. In the UNet-like module, Mamba-based encoder aims to capture long-range correspondence while the CNN-based decoder is for local feature.

\subsection{Contrastive feature learning}
From our perspectives, the frameworks of VoxelMorph and its followers can be seemed as registration modules merely without effective feature extraction. These frameworks concatenate two volumes first, treat them as a single volume and establish spatial correspondence. When registering two volumes of single modality, they may work well due to the similar appearance. When it comes to multi-modality registration, it's necessary to extract similar feature of the same region from two volumes with considerably different appearance. Besides, considering that deformable registration is a pixel-wise task, we propose a fine-grained UNet with only one down sampling as our feature extractor, as shown in Fig.~\ref{fig1}.

\subsection{SynthRAD Registration Dataset}
SynthRAD 2023 dataset was proposed at a synthetic CT (sCT) campaign originally. It contains 180 accessible and rigidly registered MR-CT pairs and each pair is from the same subject. Notably, the majority of patients took both MR and CT within a day and the maximum interval between two volumes was 10 days (only one subject). Because of short shoot interval and rigid registration, we assume that two volumes from the same subject are of well spatial alignment [Citation MR-CT]. Thus, brain segmentation labels of CT are considered to be identical to that of its corresponding MR. In addition, all the voxels were resampled to 1$\times$1$\times$1 $mm^{3}$ in SynthRad 2023.  We repropose our SR-Reg dataset from SynthRad 2023.

\begin{figure}
\includegraphics[width=\textwidth]{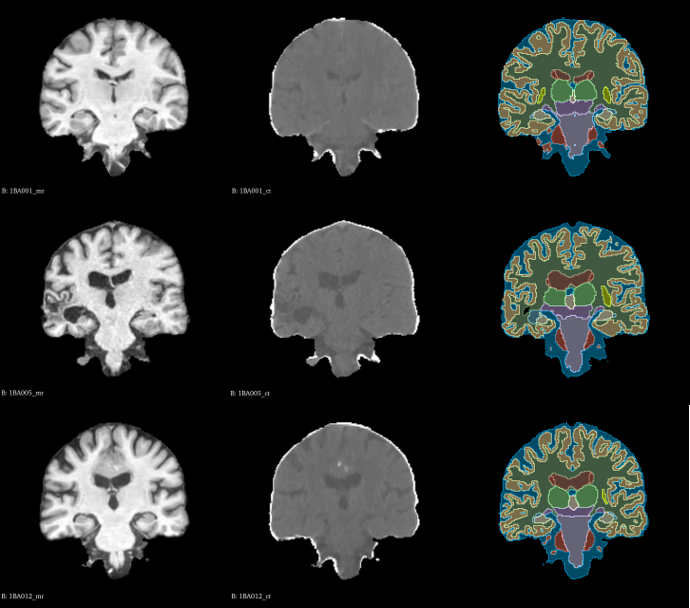}
\caption{SR-Reg Dataset. The three columns of samples are MR, CT and their corresponding segmentation respectively. Notably, volumes from each row are from the same subject.} \label{fig2}
\end{figure}

\section{Experiments}
\subsection{Experiment setting}
\subsubsection{Data usage}
We compare deformable registration methods on our multi-modality SR-Reg dataset and use 150/10/20 subjects for training, validation and test. The volumes are of 192$\times$208$\times$176 voxels at 1$\times$1$\times$1 $mm^{3}$ resolution. In our setting, moving and fixed volumes are MR and CT respectively. It's noticed that the volume pairs in test set are unchanging and the intensity of two modalities is normalized to [0,1] via min-max normalization.

\subsubsection{Model selection and implementation}
To validate our MambaMorph's superiority in multi-modal deformable registration, we compare it with two representative DL-based methods (i.e. VoxelMorph and TransMorph). 
In quantitative comparison (Tab.~\ref{tab1}), we evaluate the complete MambaMorph and the original MambaMorph (MambaMorph$_{ori}$). Specifically, MambaMorph is equipped with Mamba-based registration module, feature extractor and integration layer. For fair comparison, MambaMorph$_{ori}$ is almost identical to TransMorph, which only replaces Swin Transformer block with Mamba block. In ablation study (Tab.~\ref{tab2}), we prove the advantage of introducing fine-grained but simple feature extractor to the three registration modules. The feature extractor is a 2-layer depth UNet whose channel numbers are 16 at all layers. Notably, the MambaMorph without feature extractor includes integration layer and the MambaMorph with feature extractor is the same to the complete MambaMorph at Tab.~\ref{tab1}.

Due to the size issue, we reduce the depth of UNet (or UNet-like) network to 3 layers. The hyperparameters of $\mathcal{L}_{CL}$ and $\mathcal{L}_{smooth}$ are set to 0.001 and 0.1 respectively and the learning rate is 0.0001. Each model is trained with Adam optimizer for 100 epochs. All the experiments are implemented by PyTorch on NVIDIA GeForce RTX 4090 GPUs.

\subsubsection{Metrics}
We use the mean dice similarity coefficients (Dice) of the regions of interest [note] and 95$\%$ Hausdorff distance (HD95) to measure the registration accuracy, percentage of negative Jacobian matrix ($|J_{\phi}| \leq 0$) to evaluate the diffeomorphic property of the deformation field, inference time (Time), GPU memory usage (Memory) and amount of parameter (Param.) to evaluate the practicality of model.  

\subsection{Results and Analysis}
\subsubsection{Quantitative comparison}
From Tab.~\ref{tab1}, we can see that our MambaMorph$_{ori}$ surpasses TransMorph in nearly all aspects except for diffeomorphism. Only with a single substitution, can our registration module be more accurate, more lightweight and faster than TransMorph. These experiments demonstrate the excellence of Mamba in contrast to Transformer, which may attribute to Mamba's outstanding long-range modeling ability and its nearly linear complexity. Thanks to the efficient Mamba block, our MambaMorph is about 8\% higher on Dice and 0.7 lower on HD95 than TransMorph, consuming more inference time and memory slightly. Due to superior registration performance, acceptable inference time and low 
memory consumption, we assume that Mamba block is more suitable than Transformer in practical deformable registration. 

\begin{table}
\caption{Registration performance comparison on SR-Reg dataset. The units of Dice, HD95, $|J_{\phi}| \leq 0$, Time, Memory, Param. are $\%$, mm, $\%$, second, Gb and Mb respectively. Methods of bold font are ours, values of bold font are the best.}\label{tab1}
\begin{tabular}{|l|l|l|l|l|l|l|}
\hline
Method &  Dice $\uparrow$ & HD95 $\downarrow$ & $|J_{\phi}| \leq 0$ $\downarrow$ & Time $\downarrow$ & Memory $\downarrow$ & Param.\\
\hline
Affine initialization &  62.42$\pm$3.29 & 3.73$\pm$0.41 & - & - & - & -\\
VoxelMorph\cite{voxelmorph} & 70.88$\pm$2.99 & 3.15$\pm$0.37 & {\bfseries0.17$\pm$0.01} & {\bfseries0.05} & {\bfseries2.58} & 0.09\\
TransMorph\cite{transmorph} &  75.08$\pm$2.20 & 2.76$\pm$0.35 & 0.36$\pm$0.04 & 0.16 & 6.15 & 14.29\\
{\bfseries MambaMorhph}$_{ori}$ & 77.84$\pm$1.91 & 2.54$\pm$0.35 & 0.73$\pm$0.03 & 0.15 & 5.32 & 7.31\\
{\bfseries MambaMorhph} & {\bfseries82.71$\pm$1.45} & {\bfseries2.00$\pm$0.22} & 0.34$\pm$0.02 & 0.27 & 7.60 & 7.59\\
\hline
\end{tabular}
\label{tab1}
\end{table}

\subsubsection{Ablation study}
Through the ablation study at Tab.~\ref{tab2}, we prove the significance of feature learning in multi-modality deformable registration. Comparing the methods without feature extractor, registration modules together with feature extractor increase by at least 3\% in terms of Dice. We would like to mention that the feature extractor is only a 2-layer UNet with 16 channels, so it doesn't bring too much computation burden. Only a simple feature extractor can make such a considerable improvement, we believe that it's owed to feature learning. The former registration frameworks concatenate moving and fixed volumes at the first stage instead of extracting their own feature respectively. In other words, they treat two volumes as a single one and try to capture spatial correspondence directly. When registering volumes of single modality, the scheme work well due to the similar appearance between two volumes. While facing multi-modality data with drastically different intensity distribution (e.g. MR-CT in this paper), the scheme might not function well enough. Equipped with a feature extractor, the registration model is able to project volume intensity into high-dimensional space in fine grain and learn effective feature which is beneficial for pixel-wise registration.

\begin{table}
\caption{Ablation study for feature extractor. The units of Dice, Time, Memory, Param. are $\%$, second, Gb and Mb respectively. Methods of bold font are ours, values of bold font are the best.}\label{tab2}
\begin{tabular}{|l|l|l|l|l|l|}
\hline
Method & Feature extractor & Dice $\uparrow$  & Time $\downarrow$ & Memory $\downarrow$ & Param.\\
\hline
VoxelMorph \cite{voxelmorph} & w/o & 70.88$\pm$2.99 &  {\bfseries0.05} & {\bfseries2.58} & 0.09\\
VoxelMorph & w/ & 76.94$\pm$1.67 &  0.15 & 4.33 & 0.16 \\
TransMorph \cite{transmorph} &  w/o & 75.08$\pm$2.20 & 0.16 & 6.15 & 14.29 \\
TransMorph &  w/ & 80.60$\pm$1.44 & 0.27 & 8.34 & 14.57 \\
{\bfseries MambaMorhph} & w/o & 78.98$\pm$2.02  & 0.16 & 5.41 & 7.31\\
{\bfseries MambaMorhph} & w/ & {\bfseries82.71$\pm$1.45} & 0.27 & 7.60 & 7.59\\
\hline
\end{tabular}
\label{tab2}
\end{table}

\section{Conclusion}
In this paper, we propose a Mamba-based multi-modality deformable registration network, called MambaMorph. Its registration module consists of Mamba block, which shows superiority over Swin Transformer block in terms of long-range modeling ability. Furthermore, we are aware of the importance of feature learning in multi-modality deformable registration and introduce a simple feature extractor for MambaMorph to learn effective and fine-grained feature. The experiments demonstrate the potential of Mamba and emphasize the significance of feature learning in multi-modality deformable registration. As far as we know, this is the first method that introduces Mamba into deformable registration. We are looking forward to incorporating this approach into clinical practice.

%
%

\bibliographystyle{splncs04}
\bibliography{mybibliography}

\end{document}